\documentclass[a4paper,twoside]{article}

\usepackage{epsfig}
\usepackage{calc}
\usepackage{amssymb}
\usepackage{amstext}
\usepackage{amsmath}
\usepackage{amsthm}
\usepackage{multicol}
\usepackage{pslatex}
\usepackage{graphicx}
\usepackage{hyperref}
\usepackage{caption}
\usepackage{subcaption}
\usepackage{booktabs}
\usepackage{makecell}
\usepackage{natbib}
\usepackage{multirow}
\usepackage{comment}
\usepackage{SCITEPRESS}     
\usepackage{float}
\floatstyle{plaintop}
\restylefloat{table}

\usepackage{titlesec}
\titlespacing\section{0pt}{16pt plus 4pt minus 2pt}{10pt plus 4pt minus 2pt}

\graphicspath{ {images/} }

\begin{document}

\title{An Assessment of the Impact of OCR Noise on Language Models}

\author{\authorname{Konstantin Todorov\orcidAuthor{0000-0002-7445-4676} and Giovanni Colavizza\orcidAuthor{0000-0002-9806-084X}}
\affiliation{Institute for Logic, Language and Computation (ILLC), University of Amsterdam, The Netherlands}
\email{kztodorov@outlook.com, g.colavizza@uva.nl}
}


\abstract{Neural language models are the backbone of modern-day natural language processing applications. Their use on textual heritage collections which have undergone Optical Character Recognition (OCR) is therefore also increasing. Nevertheless, our understanding of the impact OCR noise could have on language models is still limited. We perform an assessment of the impact OCR noise has on a variety of language models, using data in Dutch, English, French and German. We find that OCR noise poses a significant obstacle to language modelling, with language models increasingly diverging from their noiseless targets as OCR quality lowers. In the presence of small corpora, simpler models including PPMI and Word2Vec consistently outperform transformer-based models in this respect.}

\onecolumn \maketitle \normalsize \setcounter{footnote}{0} \vfill

\section{\uppercase{Introduction}}

Statistical neural language models have become the backbone of modern-day natural language processing (NLP) applications. They have proven high capabilities in learning complex linguistic features and for transferable, multi-purpose adaptability~\cite{qiu_pre-trained_2020}, in particular with the recent success of contextual models like BERT~\cite{devlin_bert_2019}. Language models' main objective is to assign probabilities to sequences of linguistic units, such as sentences made of words, possibly benefiting from auxiliary tasks. The success of neural language models has fostered a significant amount of work on understanding how they work internally~\cite{rogers_primer_2020}. In Digital/Computational Humanities, language models are primarily used as components of NLP architectures and to perform well-posed tasks. Examples include Handwritten/Optical Character Recognition (H/OCR)~\cite{kahle_transkribus_2017}, Named Entity Recognition (NER) and linkage~\cite{arampatzis_overview_2020}, modelling semantic change~\cite{shoemark_room_2019}, annotating historical corpora~\cite{coll_ardanuy_living_2020}, translating heritage metadata~\cite{banar_transfer_2020}, and many more. 

An open challenge for neural language models are low-resource settings: languages or tasks where language data is comparatively scarce, and where annotations are few~\cite{hedderich_survey_2021}. This is a known issue for many underrepresented languages, including when a language is distinctively appropriated for example via dialects and idiomatic expressions~\citet{Nguyen16compsocio}. Consequences include the possible exclusion of speakers of low-resource languages, cognitive and societal biases, the reduction of linguistic diversity and often poor generalization~\cite{ruder2020beyondenglish}. Historical language data are also comparatively less abundant and sparser than modern-day data~\cite{Piotrowski_2012,Ehrmann_Colavizza_Rochat_Kaplan_2016}. \par
What is more, historical language data often poses two additional challenges: noise and variation. Noise comes from errors which should be corrected and their impact mitigated, variation instead is a characteristic of language which may constitute a useful signal. Modern-day NLP methods, including language models, `overcome' noise by sheer data size -- yet noise can still remain a problem -- and often flatten-out linguistic variation. Therefore, when employed on real-world applications in low-resource settings, or when applied out-of-domain, these methods might fall short and, crucially, they might fail to appropriately deal with noise and variation.

In this work, we contribute to bridge this gap by posing the following question: \textit{what is the impact on language models of textual noise caused by OCR?}

While recent work has focused on the impact of OCR noise on downstream tasks~\cite{hill_quantifying_2019,van_strien_assessing_2020,todorov_transfer_2020}, less is known about language models in this respect. We therefore proceed by considering a most basic empirical setting, expanding from~\cite{van_strien_assessing_2020}. First, we use data in multiple languages available in two versions: as ground truth, assumed correct and verified by humans, and as OCRed texts, containing varying degrees of noise from automated text extraction. The data we use is small when compared to modern-day language modelling standards, yet of realistic size in a Digital/Computational Humanities setting. Furthermore, we consider language models trained from scratch and do not cover fine-tuning or language model adaptation here. Lastly, for each language model under consideration we compare the results of two identically-configured language models, trained on these two versions of the same corpus, by inspecting how similar their learned vector spaces are upon convergence. In this way, we assume no universal baseline, but instead compare language models independently.

While transfer learning on language models is of extreme importance to model NLP applications~\cite{ruder_neural_2019}, we do not consider it here. The reason is the difficulty in establishing a consistent comparison. While comparing the vector spaces of language models trained from scratch is possible upon convergence, this is more problematic for fine tuning since it is difficult to know when a fine tuned model has actually converged to an accurate model of the new domain. This issue is particularly severe when using small datasets to fine tune. In fact, in the evaluation setting described above, the best result would be achieved by performing no fine tuning at all. Indeed, in this case, both the ground truth and the OCR models would be perfectly identical. A way around is to use extrinsic evaluation, and consider (the similarity in performance on) downstream tasks instead. While extrinsic evaluation has its merits, it does not allow to perform a direct assessment of a language model, but only one limited to its usefulness for downstream tasks. We therefore consider extrinsic evaluation to be complementary to the approach we pursue here.

\section{\uppercase{Related work}}

\paragraph{Neural Language Models} Vector representations of linguistic units, referred to as embeddings, have been instrumental in the success of modern neural NLP. On the one hand, as statistical models of language when trained on unsupervised objectives and, on the other hand, as components in larger NLP architectures~\cite{xia_which_2020,qiu_pre-trained_2020}. Very popular models include Word2Vec~\cite{NIPS2013_9aa42b31,DBLP:journals/corr/abs-1301-3781} and BERT~\cite{devlin_bert_2019}. A common theme of neural language modelling research over time seems to be that increasing larger parameters and datasets lead to better results~\cite{brown_language_2020,raffel_exploring_2020}. More recently, attention is increasing for low-resource languages which do not yet possess the amount of data or resources which are readily available for, say, English~\cite{ruder2020beyondenglish,hedderich_survey_2021}. As a consequence, promising work is emerging on effective small language models~\cite{schick_its_2021}.

\paragraph{Language Models and Noise.} A challenge in language modelling which is often -- yet not necessarily -- occurring in low-resource settings is noise. We can consider noise as unwanted errors in the texts, introduced by processing steps. Examples include errors in audio to text recognition or in transcription. Noise can be caused by humans, machines, communication channels; it can be systematic or not. Substantial work on noise and language models has so far focused on robustness to adversarial attacks~\cite{pruthi-etal-2019-combating}. Some approaches to language modelling can indeed be more resilient to noise than others, despite often having been introduced for other reasons (primarily dealing with out-of-vocabulary words and being language agnostic). BERT, for example, has been proven sensitive to (non-adversarial) human noise~\cite{sun2020advbert,kumar_noisy_2020}. Examples of models that can be more resilient to noise include typological language models~\cite{gerz_relation_2018,ponti_modeling_2019}, sub-word or character-level language models~\cite{kim_character-aware_2016,zhu_importance_2019,Ma_2020}, byte-pair encoding~\cite{sennrich_neural_2016}, and their extension in recent tokenization-free models~\citep{heinzerling_bpemb_2018,clark2021canine,xue2021byt5}, yet their use as noise-resilient language models remains to be fully assessed.

\paragraph{Assessing the Impact of OCR Noise.} A growing body of work is focused on assessing and mitigating the impact of OCR noise. An area of active work is that of post-OCR correction, which attempts to automatically improve on a noisy text after OCR~\cite{hamalainen-hengchen-2019-paft,boros_alleviating_2020,nguyen_neural_2020,todorov_transfer_2020}. Several recent contributions have assessed the impact of OCR noise using extrinsic evaluation and considering a variety of downstream tasks, including topic modelling, text classification, named entity recognition and information retrieval, among others~\cite{hamdi_analysis_2019,hill_quantifying_2019,van_strien_assessing_2020,boros_alleviating_2020}. While more systematic comparisons are needed, the general trend seems to indicate that OCR texts are often `good enough' to be used for downstream tasks.

It is our intent to contribute to this growing body of work by assessing the impact of OCR noise on a selection of mainstream language models, in view of informing their use and future development.

\section{\uppercase{Experimental setup}}

In this section we introduce the language models which we consider for this study and present the data we used for our experiments. Lastly, we detail the evaluation procedure to assess their resilience to OCR noise.

\subsection{Language Models}


There are several popular techniques for language modelling. Modern-day language models are typically vector-based: they map linguistic units (e.g., tokens) to vectors of a given size. How these vectors are updated and learned during training depends on the model at hand and the task(s) it focuses on. In order to compare different and popular approaches, in this study we consider Word2Vec, namely Skip-Gram with Negative Sampling (SGNS) and Continuous Bag-of-words (CBOW)~\cite{NIPS2013_9aa42b31,DBLP:journals/corr/abs-1301-3781}, and GloVe~\cite{pennington2014glove}. Furthermore, we consider  attention-based models able to make use of the context of an occurrence at inference time and not just at training time, namely BERT~\cite{devlin_bert_2019} and ALBERT~\cite{lan_albert_2020} (the latter a very fast variant of the former). Finally, we also include a language model based on co-occurrence counts re-weighted using Positive Pointwise Mutual Information (PPMI), as a baseline~\cite{Church_Hanks_1990}. 

\paragraph{PMI.} is a measure of association that quantifies the likelihood of a co-occurrence of two words taking into account the independent frequency of the two words in the corpus. Positive PMI (PPMI) leaves out negative values, under the assumption that they might capture unreliable and therefore uninformative statistics about the word pair. Formally, PMI and PPMI are calculated as follows:
\begin{align}
    \text{PMI}(w,c) &= \text{log}_2 \frac{P(w,c)}{P(w)P(c)} \\
    \text{PPMI}(w,c) &= \text{max} \left(\text{PMI}(w,c), 0 \right)
\end{align}

Where $P(w,c)$ is the probability of the word $w$ occurring together with word $c$, and the denominator contains the individual probabilities of these words occurring independently.

\paragraph{Skip-gram.} positions each word in a vector space such that similar words are closer to each other. This is achieved by using the self-supervised task of predicting the \emph{context words} given a specific \emph{target word}. To improve the speed of convergence and ease computation, we use \emph{negative sampling}. If we consider the original target and context words as positive examples, we sample negative ones at random from the corpus during training. 

\paragraph{CBOW.} uses the mirror task compared to Skip-gram, by predicting the target word using context words. For both models, the amount of context words is configurable and called \emph{window size} of the model.

\paragraph{GloVe.} is similar to PPMI, in that training is performed on the aggregated word-to-word co-occurrence matrix from a corpus, following the intuition that these encode a form of meaning.

\paragraph{BERT.} stands for Bidirectional Encoder Representations for Transformers that, at the time of introduction, gave state-of-the-art results on a wide variety of tasks. BERT is a multi-layered bi-directional transformer encoder. It uses self-supervised tasks such as masked language modelling and next sentence prediction. We apply the former to train our BERT model. The original model uses sub-word tokenization and it generates contextualised word representation at inference time.

\paragraph{ALBERT.} uses most of BERT's design choices. However, this model differs by introducing two parameter-reduction techniques that lower memory consumption and drastically decrease training time.\\

We implement the overall pipeline for our experiments and \emph{PPMI} ourselves. For \emph{CBOW} and \emph{SGNS} we use a popular Python library, \emph{Gensim}~\cite{Rehurek10softwareframework}, while for \emph{BERT} and \emph{ALBERT} we rely on \emph{HuggingFace}~\cite{huggingface}. We remark again that we always do training from scratch, without using any pre-trained model.\\

For all models, we use standard default for the used architecture tokenization. For Skip-gram, CBOW and GloVe we use simple rules, such as cleaning digits, punctuation and multiple white-spaces. For BERT and ALBERT we use Byte-level BPE tokenizer, as introduced by OpenAI and which works on sub-word level \citep{wang2020neural}. For these models, we additionally split the data at 500 characters due to the design limitations in the original implementation. Finally, we use default configurations for the transformers, namely a \emph{vocabulary size} of 30,522 for BERT and of 30,000 for ALBERT, and an \emph{vector size} equal to 768. For ALBERT, we set the number of hidden layers and attention heads to 12 and the intermediate size to 3072 so that these are equal to their counterparts in BERT.

\subsection{Data}

We make use of the datasets provided by the International Conference on Document Analysis and Recognition (ICDAR) 2017~\cite{icdar_2017} and 2019~\cite{icdar_2019} competitions on Post-OCR text correction. These two datasets combined include ten European languages of which we consider four in this study. The data is provided in three versions: OCR, aligned OCR and ground-truth. We make use of the aligned OCR and ground-truth versions for the purpose of this study and combine the training and evaluation sets together.

\begin{table*}[t]
    \centering
    \begin{tabular}{l|rrr|rrr|r}
        \toprule
        &
        \multicolumn{3}{c}{\textbf{Documents}} & \multicolumn{3}{c}{\textbf{Characters per doc.}} &
        \textbf{Total} \\
        & Total & Aligned (\% of total) & Split &
          Avg & Min & Max & 
          \textbf{characters} \\
        \midrule
        Dutch & 150 & 149 (99.3\%) & 5346 & 4593 & 42 & 16,028 & 
        688,934 \\
        English & 963 & 951 (98.8\%) & 51,689 & 6866 & 2 & 869,953 & 
        6,612,108 \\
        French & 3993 & 3616 (90.6\%) & 84,676 & 2660 & 2 & 195,177 & 
        10,620,966 \\
        German & 10,032 & 1738 (17.3\%) & 128,662 & 1581 & 126 & 16,187 & 
        15,856,445 \\
        \bottomrule
    \end{tabular}
    \caption{Dataset statistics calculated on ground truth corpora versions. The column \textit{Aligned} reports the number of documents where OCR versions and ground truth are perfectly aligned. The column \emph{Split} reports the number of documents after document splitting for transformer-based models, which require equally-sized data points.}
    \label{tab:data-stats}
\end{table*}

From Table~\ref{tab:data-stats} we show how different the four languages are in terms of corpus size. Dutch and English languages contain comparatively fewer documents but of longer average size, while French and German have more, usually shorter documents. In order to assess whether having more data for languages with a smaller corpus would alter our results, we experimented with adding data from the National Library of Australia's Trove newspaper archive~\footnote{\hyperlink{https://trove.nla.gov.au}{https://trove.nla.gov.au}.} to the English corpus, but discovered that it does not lead to any differences in the outcomes of our experiments when tested on several of our configurations. We therefore leave this out and only report results using ICDAR 2017+2019 data in what follows.

\begin{figure}
    \begin{subfigure}{.475\textwidth}
        \centering
        \includegraphics[width=1\textwidth]{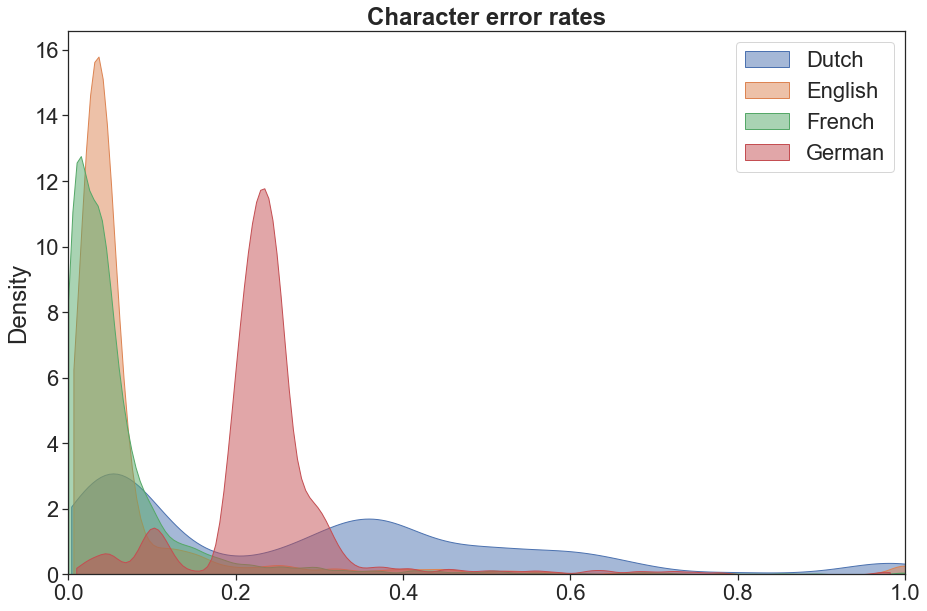}
        \captionof{figure}{Character-level}
        \label{fig:character-error-rates}
    \end{subfigure}
    \hfill
    \par\medskip
    \begin{subfigure}{.47\textwidth}
        \centering
        \includegraphics[width=1\textwidth]{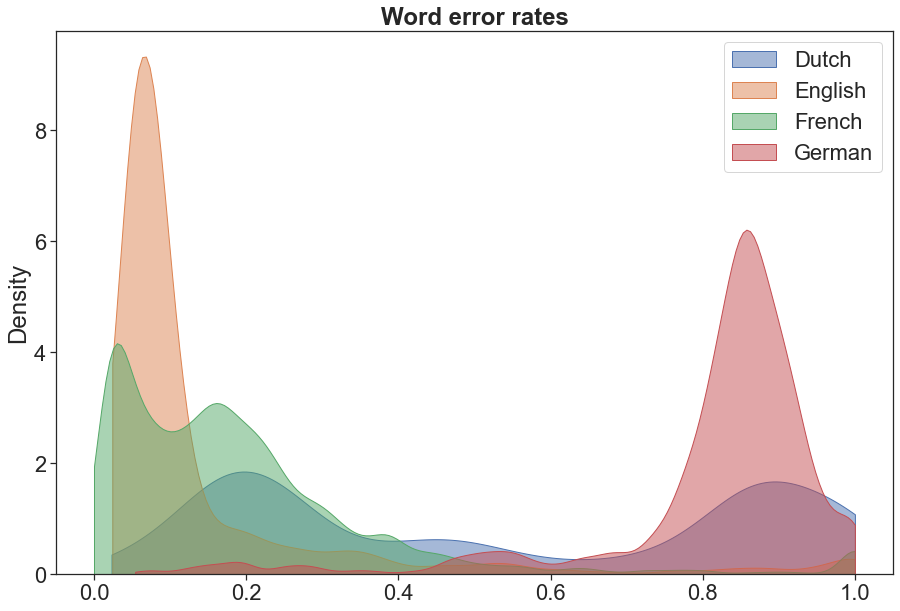}
        \captionof{figure}{Word-level}
        \label{fig:word-error-rates}
    \end{subfigure}
    \par\medskip
    \caption{OCR error rates per language, as distribution of document scores.}
\end{figure}

OCR error rates, per language and averaged over documents are given in Table~\ref{tab:error-rates}, alongside the distribution of character error rates in Figure~\ref{fig:character-error-rates} and of word error rates in Figure~\ref{fig:word-error-rates}. The error rates on character level are calculated by comparing characters on the same position in the OCR and aligned ground-truth versions. Word error rates are calculated following the \textit{de facto} standard approach of word errors to processed words~\cite{Morris_Maier_Green_2004}. Documents which are having misaligned OCR and ground-truth versions are excluded from the error rates calculation. It is worth noting that these represent a significant proportion for the German language (Table~\ref{tab:data-stats}).

\begin{table}[t]
    \centering
    \begin{tabular}{l|rr}
        \toprule
        \multirow{2}{*}{\textbf{Language}} & \multicolumn{2}{c}{\textbf{Error rate level}} \\
        & \textbf{Character} & \textbf{Word} \\
        \midrule
        Dutch & 0.286 & 0.536 \\
        English & 0.075 & 0.146 \\
        French & 0.064 & 0.193 \\
        German & 0.240 & 0.813 \\
        \bottomrule
    \end{tabular}
    \caption{OCR error rates per language, averaged over documents.}
    \label{tab:error-rates}
\end{table}

Before using each corpus for training, we perform the following pre-processing steps for all of our configurations \textit{except for BERT and ALBERT}. We remove numbers and punctuation from the data, lowercase all characters, substitute multiple white-spaces with one, also removing leading and trailing white-spaces in the process, and finally split the different words into tokens. We then build the vocabulary for each version of each corpus -- ending up with two versions per corpus: OCR and ground truth -- and remove all tokens that occur less than five times overall (per version). We pick this number following previous research and in order to reduce the computational requirements of training our models. For transformer-based models instead, we only split words into sub-word tokens and replace multiple white-spaces with one. These are typical pre-processing steps in view of presenting as much contextual information as possible to the model.

\subsection{Evaluation}

Our goal is to compare two versions of the same model configuration, each trained from scratch, on the OCR and the ground-truth versions of the same corpus. We remind the reader that our evaluation therefore does not provide any indication of the relative benefit of using one language model versus another in terms of their capacity to represent language or perform on downstream tasks. Rather, we assess and compare the resilience of each model to OCR noise. 

Our evaluation procedure is composed of the following steps:
\begin{enumerate} 
    \item For each corpus/language, we start by taking the intersection of the words that are part of all vocabularies/models. For transformer-based models, which do not have word-level vocabularies, we use the vocabulary intersection from the other models. 
    \item For transformer-based models, we output the hidden states for all words from the intersected vocabulary. For words which are split into multiple sub-word tokens due to BERT and ALBERT tokenization, we take the mean values. This approach leaves out the contextual information from the inferred embeddings, yet it ensures proper comparison with different architectures.
    \item For each corpus/language, model configuration, and word in the vocabulary intersection, we compute the cosine similarity with every other word in the vocabulary intersection.
    \item We then compare the amount of overlap in the top \textit{N} fraction of closest words (neighbors) in the two versions of the same corpus (OCR and ground truth). In this way, we are able to assess to what extent the two models agree on what is the neighborhood for each word in the vocabulary intersection.
    \item Since different corpora/languages possess varying vocabulary sizes, we use the percentage over the total dataset-specific vocabulary intersection. \textit{N} is thus ranging from 0.01 (1\%) to 1.0 (100\% top neighbors).
    \item Following~\cite{Gonen_Jawahar_Seddah_Goldberg_2020}, we use neighbor overlap as our main evaluation metric. That is the proportion of overlapping neighbours for any word in the vocabulary intersection, defined using the following formula:
    \begin{align}\label{eq:overlap}
        \text{overlap}@k(r_{OCR}, r_{truth}) = \frac{|r_{OCR}^k \cap r_{truth}^k|}{k} 
    \end{align}
    Where $r_{OCR}^k$ are the top-$k$ neighbors of word $r$ in the $OCR$ model, and $r_{truth}^k$ in the ground $truth$ model respectively. $k$ is taken to correspond to the top \textit{N} fraction of neighbors for each specific vocabulary intersection.
    \item For example, if we have a vocabulary intersection of size 1000 and evaluate at $N = 0.01$, we would take the $k = 10$ closest words for each word. If on average 5 out of 10 words correspond in the two versions/models, we would have an average 0.5 overlap score (Equation \ref{eq:overlap}).
\end{enumerate}

Further empirical settings are as follows. In agreement and to ensure compatibility with previous work on Word2Vec embeddings, we use embedding size of 300 for English, German and French languages and 320 for Dutch~\cite{tulkens2016evaluating}. We use the AdamW optimizer~\cite{AdamW} for transformer based models. We further assess models using two learning rates -- which we label \emph{fast} and \emph{slow} -- that are comparatively higher/lower in order to verify the effect of learning speed on our evaluation. Since BERT and ALBERT usually benefit from lower learning rates compared to simpler models such as CBOW and Skip-gram, we adjust accordingly. The learning rates which correspond to fast and slow for each model are shown in Table~\ref{tab:model-size-comparison}, they have been selected following commonly used default values in the literature.

To control for the stochasticity of the training process, we show all results by averaging three different runs for each hyper-parameter configuration and model type.

\begin{table}[h]
    \centering
    \sffamily
    \begin{tabular}{l|ccc}
        \toprule
        \multirow{2}{*}{\textbf{Model}} & \multirow{2}{*}{\shortstack{\textbf{Number of}\\\textbf{parameters}}} & \multicolumn{2}{c}{\textbf{Learning rates}} \\
        & & \textbf{Fast} & \textbf{Slow} \\
        \midrule
        PPMI & \multicolumn{3}{c}{N/A} \\
        GloVe & \multicolumn{3}{c}{N/A} \\
        \midrule
        CBOW & 1.7M & \multirow{2}{*}{$1\mathrm{e}{-3}$} & \multirow{2}{*}{$1\mathrm{e}{-4}$} \\
        Skip-Gram & 7.5M & & \\
        \midrule
        BERT & 108M & \multirow{2}{*}{$1\mathrm{e}{-4}$} & \multirow{2}{*}{$1\mathrm{e}{-5}$} \\
        ALBERT & 12M & & \\
        \bottomrule
    \end{tabular}
    \caption{Model parameter size comparison and default learning rates.}
    \label{tab:model-size-comparison}
\end{table}

\begin{figure}[t]
    \centering
    \includegraphics[width=.45\textwidth]{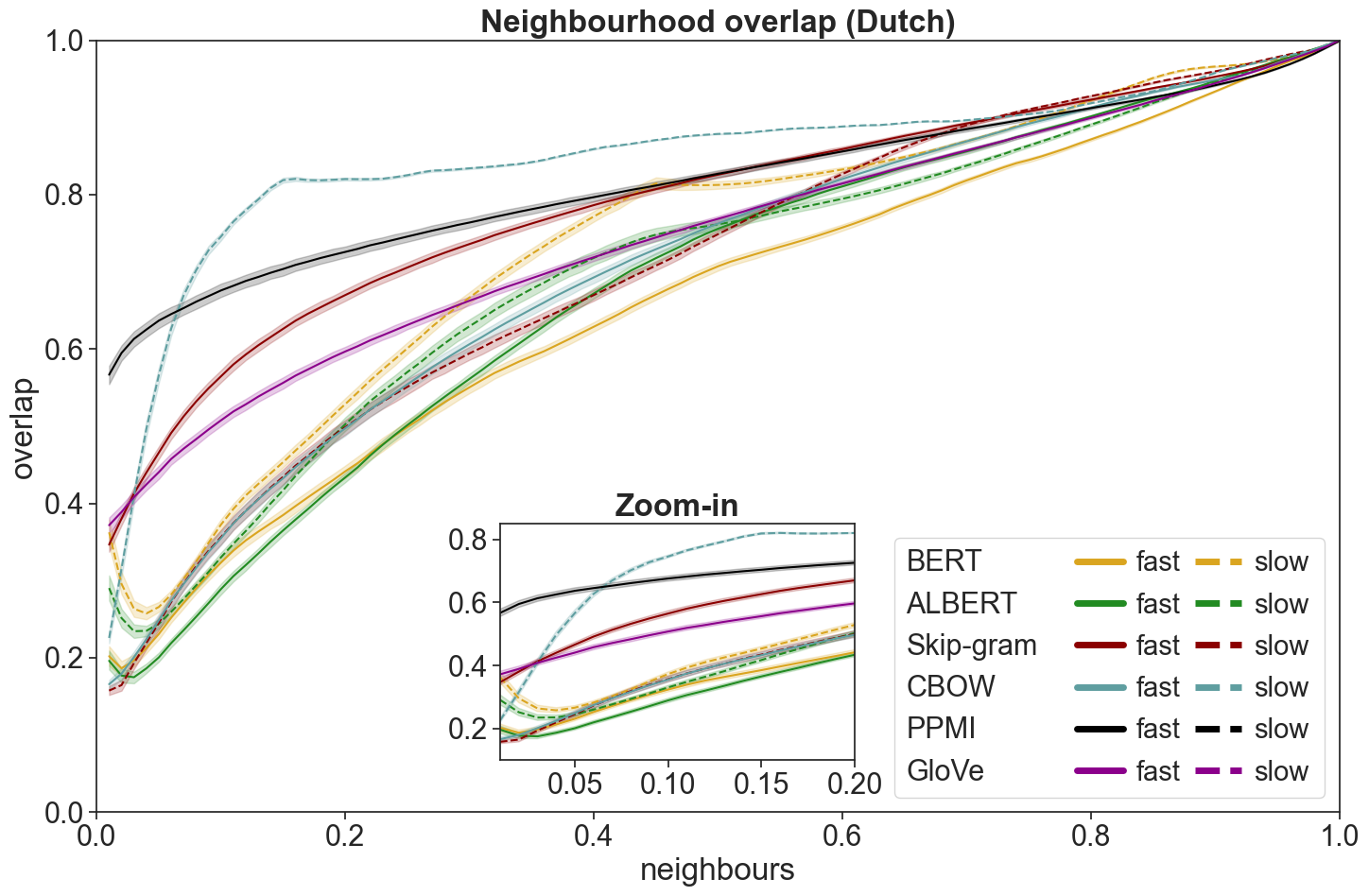}
\caption{Neighbourhood overlaps (Dutch). 95\% bootstrapped confidence intervals are provided at .01 intervals.}
    \label{fig:neighbourhood-overlaps-dutch}
\end{figure}

\begin{figure}[t]
    \centering
    \includegraphics[width=.45\textwidth]{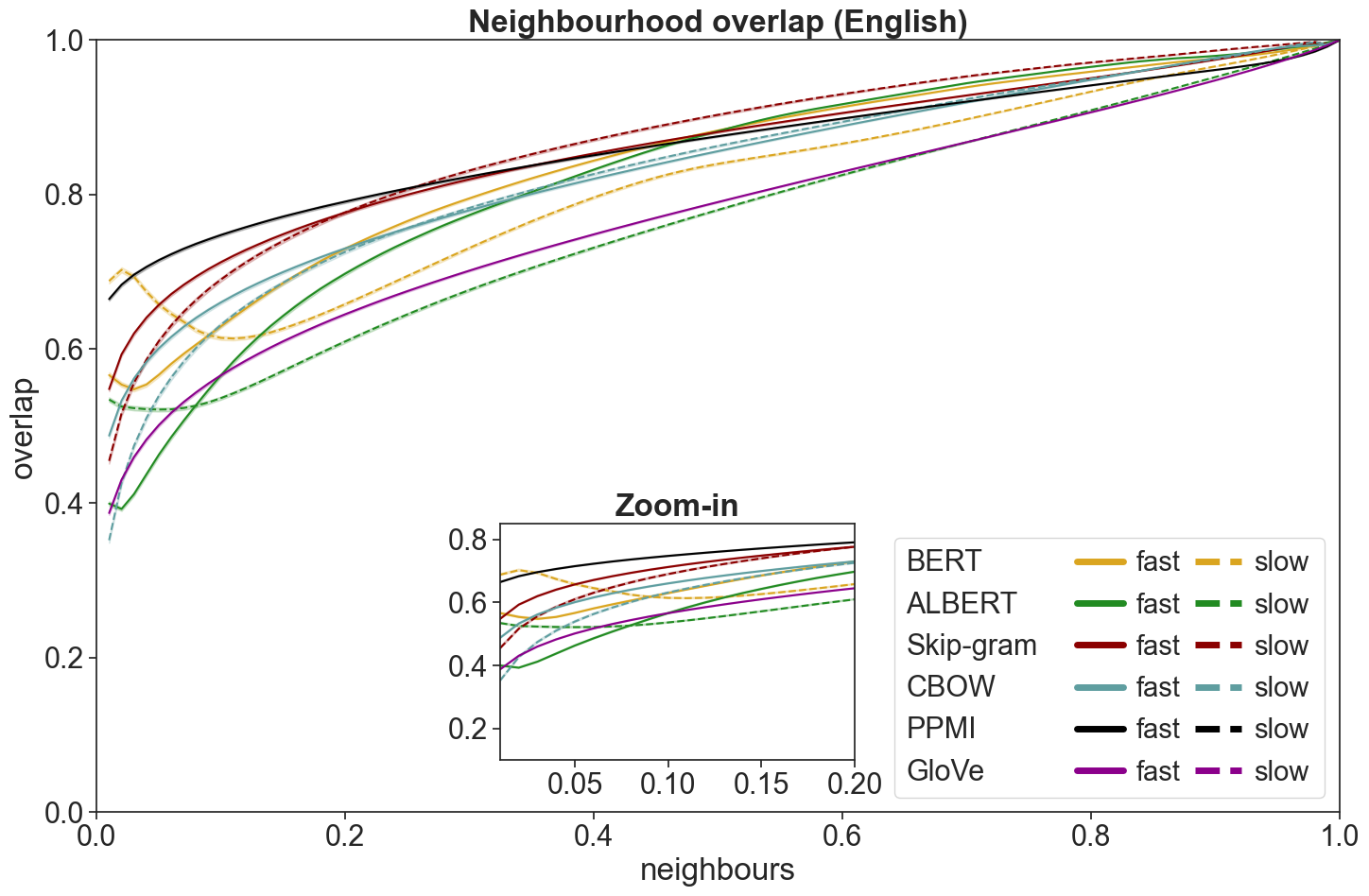}
    \caption{Neighbourhood overlaps (English). 95\% bootstrapped confidence intervals are provided at .01 intervals.}
    \label{fig:neighbourhood-overlaps-english}
\end{figure}

\begin{figure}[t]
    \centering
    \includegraphics[width=.45\textwidth]{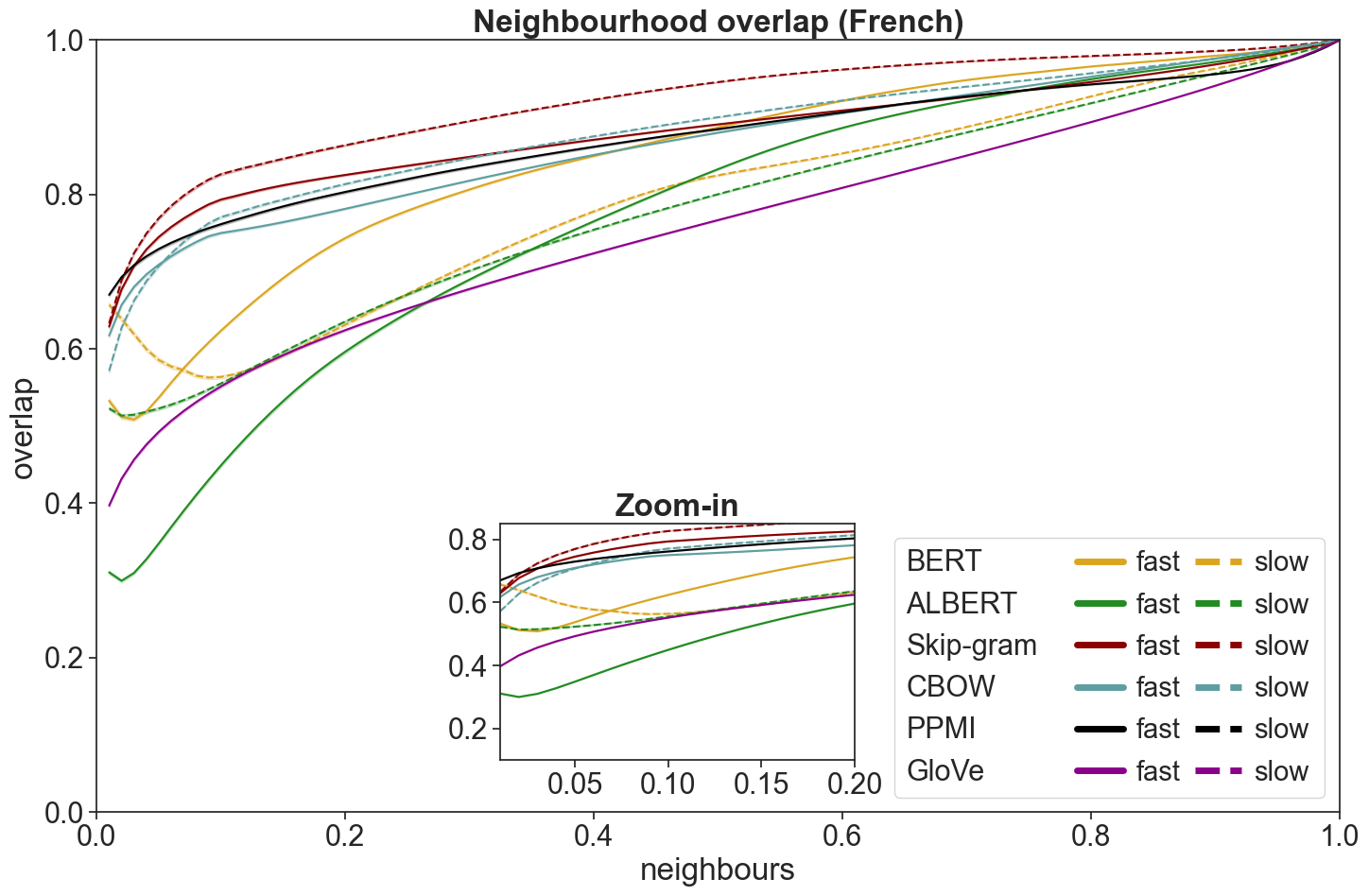}
    \caption{Neighbourhood overlaps (French). 95\% bootstrapped confidence intervals are provided at .01 intervals.}
    \label{fig:neighbourhood-overlaps-french}
\end{figure}

\begin{figure}[t]
    \centering
    \includegraphics[width=.45\textwidth]{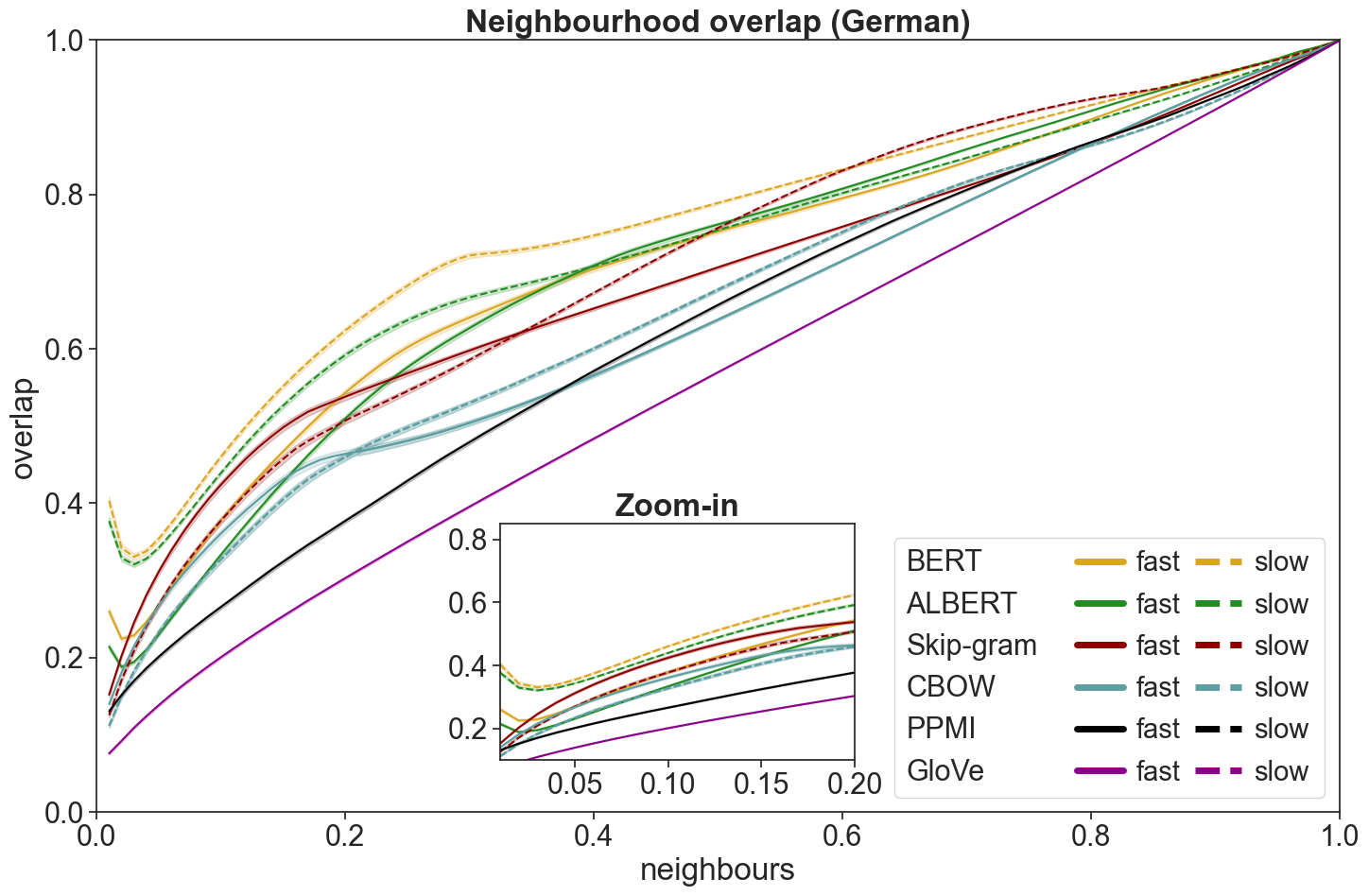}
    \caption{Neighbourhood overlaps (German). 95\% bootstrapped confidence intervals are provided at .01 intervals.}
    \label{fig:neighbourhood-overlaps-german}
\end{figure}

\section{\uppercase{Results}}

Our results are shown in a series of mirrored plots, one per language. The plots show the neighborhood overlap (Eq.~\ref{eq:overlap}) on the y axis, at varying values of $N$ from 0.01 (1\%) to 1.0 (100\%), on the x axis. Naturally, higher values of $N$ are somewhat less interesting and more trivial than lower values of $N$ (overlap of the most similar neighbors), therefore each plot also shows a zoom-in inset for values on $N$ between 0.01 (1\%) and 0.2 (20\%). Confidence intervals are added and point to the significance of all results.

Starting with Dutch, in Figure~\ref{fig:neighbourhood-overlaps-dutch}, we can appreciate how CBOW (slow learning rate) and PPMI appear to be most resilient to OCR noise, followed by Skip-gram (fast learning rate) and GloVe. Every other model configuration performs much worse at lower values of $N$. Importantly, the Dutch corpus we use is small in comparison to other corpora, and contains poor quality OCR.

Next, we show results for English in Figure~\ref{fig:neighbourhood-overlaps-english}. The English corpus has a good quality OCR and is of average size in our study. In this setting, PPMI followed by Skip-gram models perform best, notwithstanding a good performance of BERT (slow learning rate) at very low values of $N$. All other models play catch-up.

The results are quite clear for French, in Figure~\ref{fig:neighbourhood-overlaps-french}, which is a corpus of size and quality comparable to those of English. For French, Skip-gram, CBOW and PPMI models perform consistently better in resisting the impact of OCR noise. Compared, contextual BERT and ALBERT are lagging behind, along with GloVe.

Lastly, we report results for German in Figure~\ref{fig:neighbourhood-overlaps-german}. The German corpus is the largest in size, but the worst in terms of OCR quality. The low OCR quality clearly shows in the comparatively lower overlap scores overall. Surprisingly, for German the best performing models are BERT and ALBERT, in particular on a slow learning rate. It is worth noting that a fast Skip-gram configuration is also close.

Our results show clear trends. Firstly, the impact of OCR quality is overall quite significant: all models trained on OCR data diverge from ground truth. Language models trained on lower OCR quality corpora consistently reach lower overlap scores, as shown for Dutch and German. For example, the best performing models for French, which has good OCR quality, reach over 80\% overlap at low $N$ (5\%), while this overlap score is only reached at very high values for German (55\%). The effect of the size of a corpus does not seem to be significant, yet results for German, the only language where transformer-based models outperform the competition, warrant further scrutiny in this respect. In terms of language models, simpler seems to be better. PPMI, Skip-gram and CBOW models consistently perform above transformer-based models (BERT, ALBERT) in terms of resilience to OCR noise. The only exception in this respect is German, which is plagued by OCR errors but is also the largest corpus available. Furthermore, BERT consistently outperforms ALBERT, hinting at the fact that the gains in training speed come at a cost. GloVe does not appear to be particularly resilient to OCR noise either. Lastly, the variety and lack of consistency of results when using fast and slow learning rates underlines the importance of choosing the right hyper-parameters.

We end our results by highlighting a set of limitations which constitute interesting directions for future work. Firstly, our results hold within the remits of the datasets and models we focused on: using more aligned data -- which is unfortunately costly to create --, in more languages of comparable corpus sizes, and with more varied contents and OCR quality would all be useful future contributions. Furthermore, the quality of the ground truth itself, which we took here for granted, might warrant further scrutiny. Future work could also focus on hyper-parameter fine tuning in order to reach the maximum performance with any given model, as well as on assessing how many data are necessary to reach a certain desired result with a given language model. Next, as we discussed at the beginning, our results would benefit from a complementary study focused on the extrinsic evaluation of the impact of OCR on language models, in particular when used as components for machine learning architectures focused on downstream tasks. Related to this point, we decided not to consider pre-trained models: approaching the same research question via extrinsic evaluation would allow to overcome such limitation.

\section{\uppercase{Conclusion}}

We have assessed the impact of OCR noise on a variety of language models. Using data in Dutch, English, French and German, of different sizes and OCR qualities, we considered two aligned versions of the same corpus, one OCRed and one manually corrected (ground truth). Two identical instances of each language model were in turn trained from scratch over these two versions of the same corpus, and the similarity of the resulting vector spaces was assessed using word neighborhood overlap. This approach allowed us to assess and compare the resilience of each language model to OCR noise, independently.

We show that OCR noise significantly impacts language models, with a steep degradation of results for corpora with lower OCR quality. Furthermore, we show that `simpler' language models, including PPMI and the Word2Vec family (Skip-gram and CBOW) are often more resilient to OCR noise than recent transformer-based models (BERT, ALBERT). We also show that the choice of key hyper-parameters, such as the learning rate, significantly impacts results as well. The size of a corpus might also be an important factor, as suggested by transformer-based models performing best with the largest corpus available (German), yet more experiments are needed to untangle its effects.

While several limitations and opportunities for future work have been discussed, we believe the two most important next steps to be the development of multilingual evaluation corpora of similar size and OCR quality in order to study the impact of OCR noise more systematically, and the need to complement our study with an extrinsic assessment on downstream tasks making use of (possibly pre-trained) language models. Nevertheless, we have shown that OCR noise poses a significant obstacle to language modelling. In the presence of small datasets, simpler models including PPMI and Word2Vec are more resilient to OCR noise than transformer-based models trained from scratch. 

\section*{\uppercase{Code and data availability}}

All data is openly provided by the International Conference on Document Analysis and Recognition (ICDAR) 2017~\cite{icdar_2017} and 2019~\cite{icdar_2019} competitions on Post-OCR text correction.

Our code base is publicly available and described at \hyperlink{https://doi.org/10.5281/zenodo.5799211}{https://doi.org/10.5281/zenodo.5799211} \citep{konstantin_todorov_2021_5799211}.

\bibliographystyle{apalike}
{\small
\bibliography{main}}

\end{document}